\documentclass[conference]{IEEEtran}
\IEEEoverridecommandlockouts
\usepackage{cite}
\usepackage{amsmath,amssymb,amsfonts}
\usepackage{algorithmic}
\usepackage{graphicx}
\usepackage{textcomp}
\usepackage{xcolor}
\usepackage{hyperref}       
\usepackage{url} 
\usepackage{amsmath}
\usepackage{amsthm}
\usepackage{float}
\usepackage{graphicx}
\usepackage{subfigure}

\newtheorem{lemma}{Lemma}

\def\BibTeX{{\rm B\kern-.05em{\sc i\kern-.025em b}\kern-.08em
    T\kern-.1667em\lower.7ex\hbox{E}\kern-.125emX}}
\begin{document}

\title{Attention Link: An Efficient Attention-Based Low Resource Machine Translation Architecture}

\author{\IEEEauthorblockN{Zeping Min}
\IEEEauthorblockA{\textit{School of Mathematical Sciences} \\
\textit{ Peking University }\\
Beijing, China \\
zpm@pku.edu.cn}
}
\maketitle

\begin{abstract}
Transformers have achieved great success in machine translation, but transformer-based NMT models often require millions of bilingual parallel corpus for training. In this paper, we propose a novel architecture named as attention link (AL) to help improve transformer models' performance, especially in low training resources. We theoretically demonstrate the superiority of our attention link architecture in low training resources. Besides, we have done a large number of experiments, including en-de, de-en, en-fr, en-it, it-en, en-ro translation tasks on the IWSLT14 dataset as well as real low resources scene on bn-gu and gu-ta translation tasks on the CVIT PIB dataset. All the experiment results show our attention link is powerful and can lead to a significant improvement. In addition, we achieve a 37.9 BLEU score, a new sota, on the IWSLT14 de-en task by combining our attention link and other advanced methods.
\end{abstract}

\begin{IEEEkeywords}
transformer, low resource, machine translation, theoretical analysis
\end{IEEEkeywords}

\section{Introduction}
The machine translation task is one of the most important natural language processing tasks. In general, there are three paradigms for machine translation: Rule-based machine translation, Statistical machine translation, and Neural machine translation. Rule-based machine translation e.g. \cite{rajan2009rule}, \cite{hurskainen2017rule}, \cite{sghaier2020rule}, \cite{shiwen2014rule} and \cite{forcada2011apertium} is the most traditional machine translation method, which often relies on fixed rules extracted by experts to operate such as  The  statistical machine translation \cite{artetxe2018unsupervised}, \cite{och2003minimum} \cite{och1999improved} and \cite{zens2002phrase}. usually works by constructing a statistical translation model by statistically analyzing a large number of parallel corpus and then using that model for translation 
Neural Machine Translation (NMT) technology has made remarkable breakthroughs in recent years.  Before the attention mechanism, the most commonly used neural machine translation models are RNN  \cite{kalchbrenner2013recurrent}, LSTM \cite{sutskever2014sequence}, GRU \cite{cho2014learning} and there are lots of works based on them such as \cite{datta2020neural}, \cite{hermanto2015recurrent}, \cite{jian2022lstm} and \cite{xu2021multi}. Although both RNN-based and LSTM-based models have achieved great success, they face some intractable problems. For example, 
both of them struggle to translate long sentences.
\subsection{Attention based Neural machine translation}

\cite{bahdanau2014neural} try to use the attention mechanism in the machine translation task. The design of the attention mechanism is inspired by imitating human attention to quickly filter out high-value information from a large amount of information. A fully attention-based NMT model is proposed in \cite{vaswani2017attention}. It achieves excellent results on the WMT dataset, which demonstrates the amazing potential of attention mechanisms in the neural machine translation task. Since then, various NMT models based on the attention mechanism have become mainstream models in the machine translation task. For example, \cite{beltagy2020longformer},  \cite{zhu2021long}, \cite{zhang2021poolingformer} and \cite{rae2019compressive} try to  make it easier for the transformer to handle long sentences. \cite{kitaev2020reformer} and \cite{wang2020linformer} try to make the transformer train faster and require less memory. \cite{wang2019learning} try to train a deep transformer. There is no doubt that transformer models have achieved great success in machine translation, but transformer-based language models often require millions of bilingual parallel corpus to train and often require careful tuning of hyper-parameters. It is difficult to get enough parallel corpora for some specific language pairs. In this paper, we propose an easy-to-implement and effective architecture called attention link. Theoretical and experimental analyzes demonstrate the effectiveness of the attention link architecture. We believe that attention links are easy to capture more general semantic information in sentences and are robust to noise. Furthermore, our architecture can be easily deployed in transformer-based models without adding extra parameters.
 Our contributions are as below:
\begin{itemize}
\item[*] We propose attention link (AL), a novel architecture to help improve transformer models' performance, especially in low training resources. 
\item[*] We theoretically explain the superiority of attention link architecture.
\item[*] A large number of experiments have verified the effectiveness of the attention link. All the experiment results show our attention link is powerful and can lead to a significant improvement, especially under low-resource translation conditions. We also achieve a new state-of-the-art on the IWSLT14 de-en dataset.
\end{itemize}
\section{Related work}
\subsection{Improve through additional corpus}
One way to improve the performance of neural machine translation with low resources is to use the additional corpus, including the data augmentation method and other corpus which is much easier to obtain. a monolingual corpus of a language than to obtain a bilingual parallel corpus of a language pair. Pre-training  \cite{mikolov2013efficient} and back translation \cite{sennrich2015improving}, \cite{edunov2018understanding} and \cite{li2020revisiting} are two ways to use  monolingual corpus. Pre-training boosts performance by learning a good representation. For example, \cite{qi2018and}.  \cite{devlin2018bert} learning the representation through predicting mask words in sentences and it has achieved great success on numerous NLP tasks.

A classic and commonly used method is back translation \cite{sennrich2015improving}, \cite{hoang2018iterative}. The main idea of back translation \cite{sennrich2015improving} is to use the monolingual corpus to improve the performance of models. Back translation can often significantly improve machine translation performance in low-resource conditions and there are many works on back-translation methods.

\subsection{Improvement through model}
There are also a lot of works on improving the model to make the model more suitable for low-resource conditions. 

One of the easiest ways to improve the model is to choose appropriate hyper-parameters. \cite{araabi2020optimizing} and \cite{van2020optimal} gained performance by choosing appropriate transformer hyper-parameters. Also, there are works focused on improving the model itself. \cite{you2020hard} replaced all self-attention matrices in the encoder and decoder with fixed Gaussian distribution. However, the obtained model performance did not decrease significantly. \cite{raganato2020fixed} fixing the attention matrix of the Transformer encoder can bring up a 3 points improvement in BLEU score in low-resource scenarios. In this work, we also focus on the attention matrix. For low resource conditions, we propose an adaptive architecture, attention link, from the attention matrix perspective.

\section{Attention link architecture}
In this section, we first introduce our insight and the motivation for designing the attention link. Then we precisely describe the attention link architecture and its mathematical formulations.
\subsection{Motivation}
In general, a large part of the difficulty caused by low resources comes from the difficulty in capturing adequate information. Excessive training is more likely to lead to serious overfitting. In particular, when we try to train a transformer-based neural machine translation system under low-resource conditions, it may be difficult for the transformer model to extract semantic information, but a simple memory of samples. Excessive memory may cause the model to pay too much attention to the correlation between a small number of words (subwords) and a lack of global vision. The attention matrices in the transformer model reflect the model's judgment on the correlation between words (subwords). 

\begin{figure*}[h]
    \centering
	\label{PAAbefore}
	\centering
	\subfigure[Self attention matrices in encoder]{
		\begin{minipage}{14cm} 
		\centering
        \includegraphics[width=\textwidth,height=0.37\textwidth]{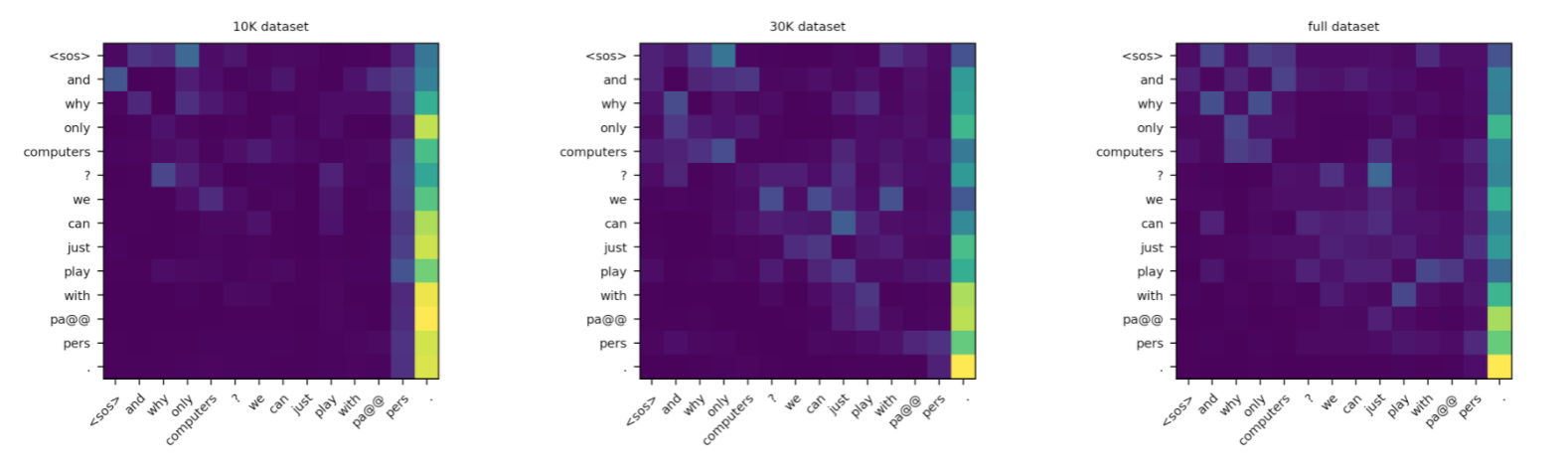} \\

		\end{minipage}
	}

	\subfigure[Cross attention matrices in decoder]{
		\begin{minipage}{14cm}
		\centering
			\includegraphics[width=\textwidth,height=0.37\textwidth]{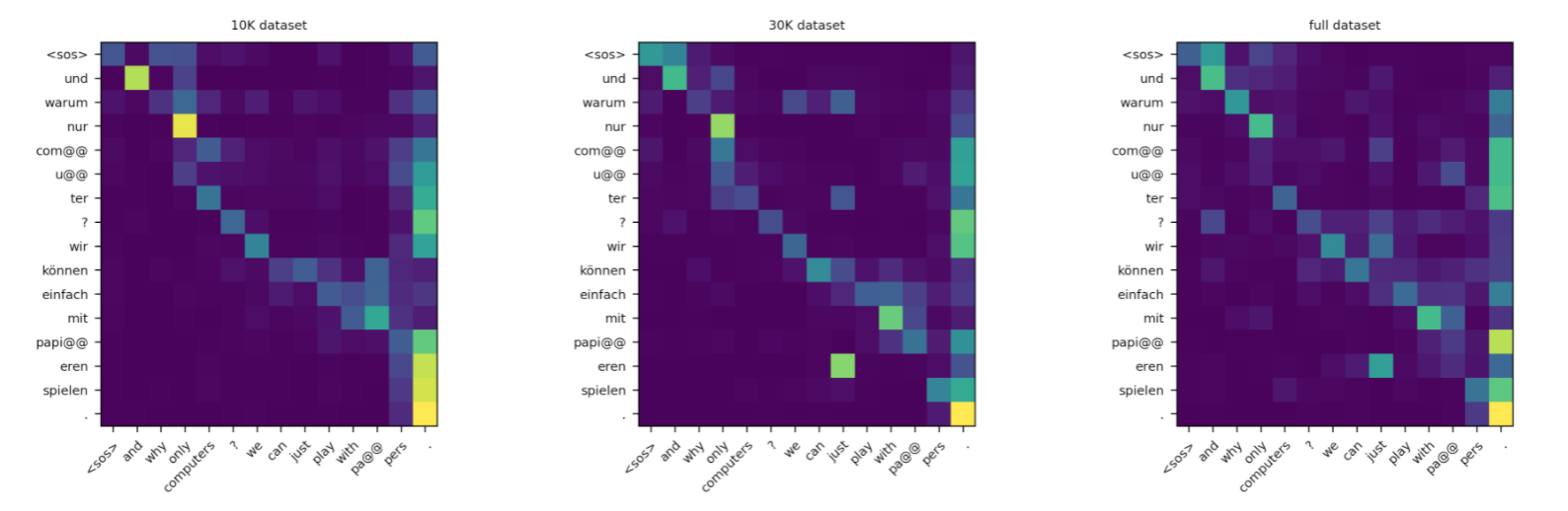} \\
		\end{minipage}
	}
	\caption{Attention matrices of transformer models trained in  different resources for en-fr translation task}
    \label{fig1}
\end{figure*}

We start with the attention matrix to verify our speculation. Firstly, we randomly select about 10K and  30K parallel corpus of en-fr IWSLT14 dataset to simulate the low resource condition  train the transformer models on the  10K, 30K training resource level, and full dataset respectively. Then we visualize the  attention matrices of transformer models trained in different resource levels on one randomly selected sentence pair in Fig. \ref{fig1}. We can observe that for the transformer model trained under low-resource conditions, both the self-attention matrix and the cross-attention matrix are significantly sparser. This means that they put most of their attention on a few word pairs, and lack a global vision. This might work well across certain sentence pairs (such as the samples memorized by the model), but the lack of a global view will cause the model to lack generalization. 

Inspired by the shortcut structure in resnet \cite{he2016deep}, we propose an attention link structure. Since the attention matrix is to capture the relationship among words, to decrease the low resource effects, we try to connect the neighbor layers' attention matrix inspired by the shortcut structure in resnet \cite{he2016deep}. Through attention links, the information between the attention matrices can be shared, thereby reducing the possibility of the single-layer attention matrix being disturbed by noise.

\subsection{Transformer and attention link}
We first briefly introduce the transformer model, specify some notation, and then give our attention link architecture.
The transformer model \cite{vaswani2017attention} is the cornerstone of many fields in recent years. The encoder and decoder of the transformer model are spliced by the encoder layer and decoder layer. Both encoder layer and decoder layer are mainly composed by $\operatorname{Self\_Attn}$ part, $\operatorname{Cross\_Attn}$ part and $\operatorname{FFN}$ part. These layers map $\boldsymbol{X} \in \mathbb{R}^{d \times n}$ to $\boldsymbol{X} \in \mathbb{R}^{d \times n}$. Mathematically, as the formulation in \cite{yun2019transformers}, we have
\begin{equation}
\begin{aligned}
   &\operatorname{Self\_Attn}_n(\boldsymbol{X}) = 
   \sum_{i=1}^{h} \boldsymbol{W}_{n,O}^{i} \boldsymbol{W}_{n,V}^{i} \boldsymbol{X} \cdot \\
   &S\left[\left(\boldsymbol{W}_{n,K}^{i} \boldsymbol{X}\right)^{T} \boldsymbol{W}_{n,Q}^{i} \boldsymbol{X}\right]
\end{aligned}
   \label{eq1}
\end{equation}

\begin{equation}
\begin{aligned}
&\operatorname{Cross\_Attn}_n(\boldsymbol{X}) = 
\sum_{i=1}^{h} \boldsymbol{W}_{n,O}^{i} \boldsymbol{V} \cdot\\ &S\left[\left(\boldsymbol{K}\right)^{T} \boldsymbol{W}_{n,Q}^{i} \boldsymbol{X}\right]
\end{aligned}
\label{eq2}
\end{equation}

\begin{equation}
\label{eq3}
\mathrm{FFN}(\boldsymbol{X})=\boldsymbol{W}_{2} \cdot \operatorname{ReLU}\left(\boldsymbol{W}_{1} \cdot (\boldsymbol{X})+\boldsymbol{b}_{1} \mathbf{1}^{T}\right)+\boldsymbol{b}_{2} \mathbf{1}^{T} 
\end{equation}
 where $\operatorname{Self\_Attn}_n$ represents the self attention part of the nth encoder layer or the nth decoder layer. $\operatorname{Cross\_Attn}_n$ represents the self attention part of  the nth decoder layer. $\boldsymbol{W}_{n,O}^{i} \in \mathbb{R}^{d \times d_{v}}, \boldsymbol{W}_{n,V}^{i} \in \mathbb{R}^{d_{v} \times d}, \boldsymbol{W}_{n,K}^{i} \in \mathbb{R}^{d_{k} \times d}, \boldsymbol{W}_{n,Q}^{i} \in \mathbb{R}^{d_{q} \times d}, \boldsymbol{W}_{2} \in \mathbb{R}^{d \times d_{hidden}}, \boldsymbol{W}_{1} \in \mathbb{R}^{d_{hidden} \times d}, \boldsymbol{b}_{2} \in \mathbb{R}^{d}, \boldsymbol{b}_{1} \in \mathbb{R}^{d_{hidden}}.$ And $ d, d_q, d_k, d_v, d_{hidden}, h $ are 6 main hyper-parameters. $d$ represents the text embedding dimension,  $d_q$ represents the query vector dimension, $d_k$ represents the key vector dimension, $d_v$ represents the value vector dimension, and $d_{hidden}$ represents the hidden layer dimension, $ h $ represents the number of transformer heads.$S$  represents the soft-max function. $K$ represents the key tensor of the last encoder. $V$ represents the value tensor of the last encoder. $S\left[\left(\boldsymbol{W}_{n,K}^{i} \boldsymbol{X}\right)^{T} \boldsymbol{W}_{n,Q}^{i} \boldsymbol{X}\right]$ represents the self attention matrix of the ith head in nth encoder layer or nth decoder layer. $S\left[\left(\boldsymbol{K}\right)^{T} \boldsymbol{W}_{n,Q}^{i} \boldsymbol{X}\right]$ represents the cross attention matrix of the ith head in nth decoder layer.

Inspired by the shortcut structure in resnet \cite{he2016deep}, our attention link structure is to connect the attention matrices between adjacent layers. Through attention links, the information between the attention matrices can be shared, thereby reducing the possibility of the single-layer attention matrix being disturbed by noise. Mathematically, we have \\

\begin{equation}
\begin{aligned}
&\operatorname{Linked\_Self\_Attn}_n(\boldsymbol{X}) =\sum_{i=1}^{h} \boldsymbol{W}_{n,O}^{i} \boldsymbol{W}_{n,V}^{i} \boldsymbol{X} \cdot \\
&S[\left(\boldsymbol{W}_{n,K}^{i} \boldsymbol{X}\right)^{T} \boldsymbol{W}_{n,Q}^{i} \boldsymbol{X}+\left(\boldsymbol{W}_{n-1,K}^{i} \boldsymbol{X}\right)^{T} \\
&\boldsymbol{W}_{n-1,Q}^{i} \boldsymbol{X} ] 
\end{aligned}
\end{equation}

\begin{equation}
\begin{aligned}
&\operatorname{Linked\_Cross\_Attn}_n(\boldsymbol{X}) =\sum_{i=1}^{h} \boldsymbol{W}_{n,O}^{i} \boldsymbol{V} \\
&\cdot S[\left(\boldsymbol{K}\right)^{T} \boldsymbol{W}_{n,Q}^{i} \boldsymbol{X}+\left(\boldsymbol{K}\right)^{T} \boldsymbol{W}_{n-1,Q}^{i} \boldsymbol{X}
] 
\end{aligned}
\end{equation}

where $\operatorname{Linked\_Self\_Attn}_n$ represents the self-attention part of the nth encoder layer or the nth decoder layer. $\operatorname{Linked\_Cross\_Attn}_n$ represents the self-attention part of the nth decoder layer. Other notations are the same as in \eqref{eq1}, \eqref{eq2}, and \eqref{eq3}.

\begin{figure*}[h]
    \centering
    
    \subfigure[Linked self attention]{
        \includegraphics[scale=0.85]{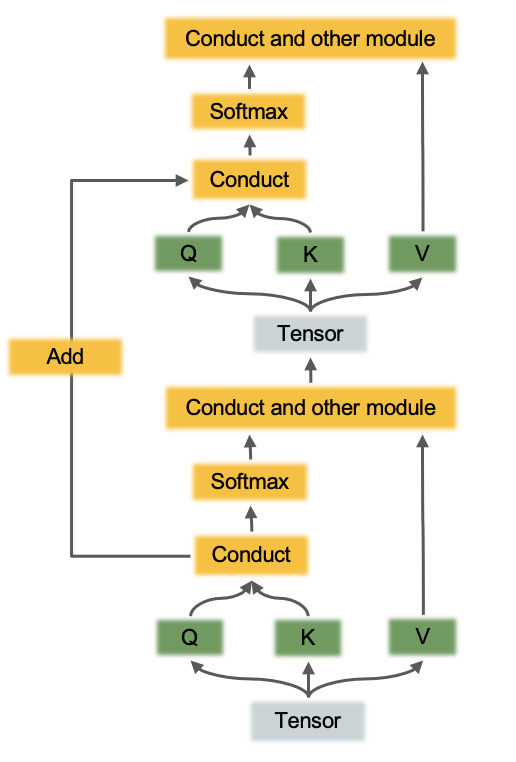}
    }
    \subfigure[Linked cross attention]{
	\includegraphics[scale=0.85]{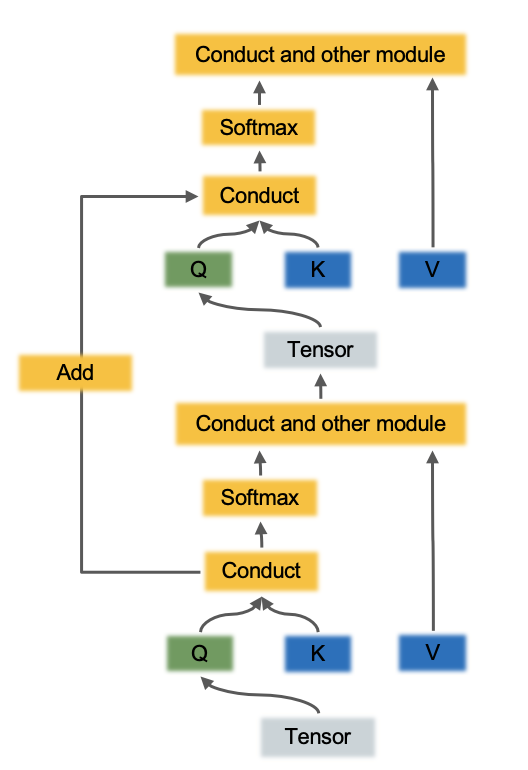}
    }
    \caption{Linked self attention and Linked cross attention}
    \label{fig.2}
\end{figure*}

The core point of the attention link is to make the attention matrix in each layer of the transformer model depend not only on the query tensor and key tensor of the current layer but also on the query tensor and key tensor of the previous layer so that each layer has a larger receptive field and at the same time alleviates the problem of low insufficient resource training biases the attention matrix. Note that the attention link architecture just replaces the $\operatorname{Self\_Attn}_n$ and $\operatorname{Cross\_Attn}_n$ parts of the encoder layer and decoder layer of the transformer model with $\operatorname{Linked\_Self\_Attn}_n$ and $\operatorname{Linked\_Cross\_Attn}_n$. This does not introduce additional parameters and is easy to implement. The structure of the attention link is shown in Fig. \ref{fig.2}. 

\section{Theory Analyze of the attention link}

\subsection{Representation ability}
We first illustrate that the attention link does not change the representation ability of the transformer model. Using $T_{h}(\theta, \cdot)$ denotes transformer model  with  hyper-parameters $h$ and trainable parameters $\theta$, $\widetilde{T}_{h}(\tilde{\theta}, \cdot)$ denotes  transformer model  with attention link as well as hyper-parameters $h$ and trainable parameters $\tilde{\theta}$. Since the attention link structure only adds the $Q, K$ product in the previous layer of the transformer model to the $Q, K$ product of the current layer, we only need the attention matrices of the last layer of $T_{h}(\theta, \cdot)$ and $\widetilde{T}_{h}(\tilde{\theta}, \cdot)$ to remain the same for both the encoder and the decoder. This can obviously be done by choosing the appropriate $Q, K$ of the last layer of the encoder or decoder in $\widetilde{T}_{h}(\tilde{\theta}, \cdot)$. In summary, we have

\begin{lemma}
 For transformer model  $T_{h}(\theta, \cdot)$ with arbitrary parameters $\theta$, there exists parameters $\tilde{\theta}$ so that the   linked transformer model  $\widetilde{T}_{h}(\tilde{\theta}, \cdot)$  equals with the $T_{h}(\theta, \cdot)$ . That is to say, 
 the representation ability of the linked transformer  will not less than the transformer.
\end{lemma}
\subsection{Robustness}
In Lemma 1, we explained that the transformer with attention link and the transformer model have the same representation ability. In this section, we try to explain the advantages of attention links in theory.
\paragraph{Notation}
We firstly give the notation in Table \ref{tab1}
\begin{table}[H]
  \caption{Notation in the analysis}
  \begin{center}
  \begin{tabular}{ll}
    \hline
    Notation  & Meaning      \\
    \hline
    $\boldsymbol{x}$     & input vector          \\
    $\boldsymbol{y}$     & (ground truth) output vector            \\
    $\tilde{\boldsymbol{y}}$ & vanilla transformer output vector   \\
    $\tilde{\tilde{\boldsymbol{y}}}$ & transformer with AL output vector \\
    $N$                  & vector length           \\
    $\boldsymbol{T}$     & (simplified) transformer operation     \\
    $P$                  & attention matrix          \\
    $\theta$             & model parameters          \\
    $\theta^{*}$         & ground truth model parameters          \\
    $\sigma(i,j)$        & error in position $(i,j)$ of $P$         \\
    $\tilde{\Delta}$     & error of $\tilde{\boldsymbol{y}}$ \\
    $\tilde{\tilde{\Delta}}$   & error of $\tilde{\tilde{\boldsymbol{y}}}$ \\
    \hline
  \end{tabular}
  \label{tab1}
  \end{center}
\end{table}
\paragraph{Analysis setting}
To simplify the analysis, we make some non-general simplifications.
Firstly we ignore the effect of pointwise linear layers and just analyze \textbf{one} layer in the transformer model. Besides, we assume both $\boldsymbol{x}$ and $\boldsymbol{y}$ are vectors with $N$ length. So mathematically we have 

\begin{equation}
\boldsymbol{T}_{\theta}(\boldsymbol{x})(j)=\sum_{i=1}^{N} \boldsymbol{x}(i) P(\theta, \boldsymbol{x}, i, j)
\end{equation}
\begin{equation}
\sum_{i=1}^{N} P(\theta, \boldsymbol{x}, i, j)=1
\end{equation}

 $P(\theta, \boldsymbol{x}, i, j)$ is attention matrix with parameters $\theta$ and $\boldsymbol{x}$ input. Note that the $P(\theta, \boldsymbol{x}, i, j)$ act as the attention matrix in the transformer model. We denote the ground truth of the parameters as $\theta^*$. So we have, 
\begin{equation}
\boldsymbol{y}(j)=\sum_{i=1}^{N} \boldsymbol{x}(i) P(\theta^{*}, \boldsymbol{x}, i, j)
\end{equation}
\paragraph{Vanilla transformer in low resource}
Since the low training resource, we can not get the exact $\theta^{*}$ as well as the $P(\theta^{*}, \boldsymbol{x}, i, j)$. We denote the error caused by low resource in $P(\theta^{*}, \boldsymbol{x}, i, j)$ as $\sigma(i, j)$ and assume that $\sigma(i, j) \stackrel{i.i.d}{\sim} N\left(0, \sigma_{0}^{2}\right)$. So mathematically we have 
\begin{equation}
\tilde{\boldsymbol{y}}(j)=\frac{1}{\widetilde{c}_{j}} \sum_{i=1}^{N} \boldsymbol{x}(i)\left(P\left(\theta^{*}, \boldsymbol{x}, i, j\right)+\sigma(i, j)\right)
\end{equation}

The $\sigma_{0}^{2}$ is a constant and $\tilde{c_j}$ is the normalization coefficient. Since the error in $P$ is relatively small, we have $\tilde{c_j} \approx 1$. 
\begin{equation}
\begin{aligned}
&\tilde{\Delta}(\tilde{\boldsymbol{y}}(j))=|\boldsymbol{y}(j)-\widetilde{\boldsymbol{y}}(j)| \approx \\
&\left|\sum_{i=1}^{N} \boldsymbol{x}(i) \sigma(i, j)\right| \leq \frac{1}{2} \sum_{i=1}^{N}\left(\boldsymbol{x}^{2}(i)+\sigma^{2}(i, j)\right)
\end{aligned}
\end{equation} 
\begin{equation}
\label{eq10}
\begin{aligned}
&\frac{1}{N} \sum_{j=1}^{N} \tilde{\Delta}(\boldsymbol{\tilde{y}}(j)) \leq\\
&\frac{1}{2 N} \sum_{i=1}^{N} \sum_{j=1}^{N} \boldsymbol{x}^{2}(i)+\frac{1}{N} \sum_{i=1}^{N} \sum_{j=1}^{N} \sigma^{2}(i, j)
\end{aligned}
\end{equation}
\paragraph{Transformer with attention link}
For the linked transformer model, since the low training resource, both previous and current layers' attention matrices also have error and we denote previous  layers' attention matrices  error as $\sigma_{pre}$. So we have $\sigma_{pre}(i, j) \stackrel{i.i.d}{\sim} N\left(0, \sigma_{0}^{2}\right)$ and $\sigma(i, j) \stackrel{i.i.d}{\sim} N\left(0, \sigma_{0}^{2}\right)$.  Due to the attention link,  we have the output $\tilde{\tilde{\boldsymbol{y}}}$:

\begin{equation}
\begin{aligned}
\tilde{\tilde{\boldsymbol{y}}}(j)=\frac{1}{\widetilde{\widetilde{c_j}}} \sum_{i=1}^{N} \boldsymbol{x}(i) (\frac{1}{2}P\left(\theta^{*}, \boldsymbol{x}, i, j\right)+\\\frac{1}{2} P_{pre}\left(\theta^{*}, \boldsymbol{x}, i, j\right)+\frac{1}{2} \sigma(i, j)+\frac{1}{2} \sigma_{pre}(i, j))
\end{aligned}
\end{equation}
If we denote $\gamma\left(\theta^{*}, \boldsymbol{x}, i, j\right))$ as 
\begin{equation}
\begin{aligned}
\gamma\left(\theta^{*}, \boldsymbol{x}, i, j\right)) = P_{pre}\left(\theta^{*}, \boldsymbol{x}, i, j\right))-P\left(\theta^{*}, \boldsymbol{x}, i, j\right) 
\end{aligned}
\end{equation}
Then we have 
\begin{equation}
\label{eq12}
\begin{aligned}
\tilde{\tilde{\boldsymbol{y}}}(j)=\frac{1}{\widetilde{\widetilde{c_j}}} \sum_{i=1}^{N} \boldsymbol{x}(i)(P\left(\theta^{*}, \boldsymbol{x}, i, j\right)+\\
\frac{1}{2} \sigma(i, j)+\frac{1}{2} \sigma_{pre}(i, j)+\frac{1}{2} \gamma\left(\theta^{*}, \boldsymbol{x}, i, j\right))
\end{aligned}
\end{equation}
Since the error in $P$ is relatively small, we also have $\tilde{\tilde{c_j}} \approx 1$.
If the ground truth $P$ difference between adjacent transformer layers  is relatively small, that is  $\gamma\left(\theta^{*}, \boldsymbol{x}, i, j\right)) \approx 0$, then  we have 
\begin{equation}
\label{eq13}
\begin{aligned}
&\tilde{\tilde{\Delta}}(\tilde{\tilde{\boldsymbol{y}}}(j))=|\boldsymbol{y}(j)-\tilde{\boldsymbol{y}}(j)|  \approx\\
&\left|\sum_{i=1}^{N}x(i)\left[\frac{1}{2} \sigma(i, j)+\frac{1}{2} \sigma_{pre}(i, j)\right]\right| \\
& \leq \frac{1}{2} \sum_{i=1}^{N}\left[\boldsymbol{x}^{2}(i)+\left(\frac{1}{2} \sigma(i, j)+\frac{1}{2}
\sigma_{pre}(i, j)\right)^{2}\right]
\end{aligned}
\end{equation}

Take the average of the elements in $\tilde{\tilde{\boldsymbol{y}}}$, we have 

\begin{equation}
\label{eq15}
\begin{aligned}
&\frac{1}{N} \sum_{j=1}^{N} \tilde{\tilde{\Delta}}(\tilde{\tilde{\boldsymbol{y}}}(j)) \leq
\frac{1}{2 N} \sum_{i=1}^{N} \sum_{j=1}^{N} \boldsymbol{x}^{2}(i)+\\
&\frac{1}{N} \sum_{i=1}^{N} \sum_{j=1}^{N}\left(\frac{1}{2} \sigma^{2}(i, j)+\frac{1}{2} \sigma_{pre}(i, j)\right)
\end{aligned}
\end{equation}

\paragraph{The superiority of attention link}
Now we compare the error of $\tilde{\boldsymbol{y}}$ and $\tilde{\tilde{\boldsymbol{y}}}$ in \eqref{eq10} and \eqref{eq15} respectively.
Note that $\sigma_{pre}(i, j) \stackrel{i.i.d}{\sim} N\left(0, \sigma_{0}^{2}\right)$ and $\frac{1}{2} \sigma(i, j)+\frac{1}{2} \sigma_{pre}(i, j)\stackrel{i.i.d}{\sim} N\left(0, \frac{1}{2}\sigma_{0}^{2}\right)$. According to the law of large numbers, for the error of $\tilde{\boldsymbol{y}}$ in \eqref{eq10} we have

\begin{equation}
\label{eq16}
\begin{aligned}
\frac{1}{N} \sum_{j=1}^{N} \sum_{i=1}^{N} \sigma^{2}(i, j) \approx E \sum_{i=1}^{N} \sigma^{2}(i, j)=N \sigma_{0}^{2}
\end{aligned}
\end{equation}

and for the error of $\tilde{\tilde{\boldsymbol{y}}}$ in \eqref{eq15}, we have
\begin{equation}
\label{eq17}
\begin{aligned}
&\frac{1}{N} \sum_{j=1}^{N} \sum_{i=1}^{N}\left(\frac{1}{2} \sigma^{2}(i, j)+\frac{1}{2} \sigma_{pre}^{ 2}(i, j)\right) \approx \\
&  E\sum_{i=1}^{N}\left(\frac{1}{2} \sigma^{2}(i, j)+\frac{1}{2} \sigma_{pre}^{ 2}(i, j)\right)=\frac{1}{2} N \sigma_{0}^{2}
\end{aligned}
\end{equation}
Comparing \eqref{eq16} and \eqref{eq17}, we find that the error output of transformer model with attention link $\tilde{\tilde{\boldsymbol{y}}}$ is reduced by $\frac{1}{2} N \sigma_{0}^{2}$ and this shows the superiority of the attention link.

\section{Experiment}
\subsection{Set up}

\subsubsection{Dataset}
We perform numerical experimental tests on six tasks of IWSLT14: en-de, de-en, en-fr, en-it, it-en, and en-ro as well as on bn-gu and gu-ta translation tasks on the CVIT PIB dataset. 

To verify the effectiveness of the attention link structure in low resource conditions, we first simulate the low resource scenes by randomly extracting about 10K and 30K parallel corpora from the full bilingual training split corpora on the en-de, de-en, en-fr, en-it, it-en, and en-ro machine translation tasks of IWSLT14 dataset. We train the models on 10K, and 30K training resource levels and full training split respectively, and evaluate the full test split for the en-de, de-en, en-fr, en-it, it-en, and en-ro translation tasks of IWSLT14. 

Then we experiment on bn-gu and gu-ta translation tasks of the CVIT PIB dataset. We randomly sample about 5\% datas of the bn-gu and gu-ta translation tasks in  CVIT PIB dataset as the test split and the rest as the training split. 

At the same time, we test the attention link structure on the full dataset of en-de, de-en, en-fr, en-it, it-en, and en-ro machine translation tasks in the IWSLT14 dataset. Besides, we combine our attention link architecture with other advanced methods and then test on the IWSLT14 en-de dataset.

\subsubsection{Model hyper-parameters setting}
We mainly compare the two structures: transformer \cite{vaswani2017attention} and transformer with attention link. We use the standard 6-layer encoder and 6-layer decoder structure.
Note that the attention link does not introduce additional parameters to the model. For the sake of fairness, the hyper-parameters of the transformer \cite{vaswani2017attention} and transformer with attention link models are kept the same in the experiments. In our experiments, we set $ d=512, d_q=128, d_k=128, d_v=128 , d_{hidden}=1024 , h=4$. Our numerical experiments are based on the fairseq\footnote{https://github.com/facebookresearch/fairseq} code. Both transformer  and transformer with attention link  are optimized using the adam \cite{kingma2014adam} method, using the default setting in fairseq  $\beta_0=0.9, \beta_1=0.98, \epsilon=10^{-8}, weight decay=10^{-4}$. Besides, we use the warmup strategy to control the change in learning rate as well as the default setting in fairseq  $warmup step=4000$ with $lr=5\times10^{-4}$ on the en-de, de-en, en-fr, en-it, it-en, and en-ro machine translation tasks in the IWSLT14 dataset.

\subsection{Result}
In the following Table \ref{n1_tab2}, we show some translation examples of the  models trained in  the 10K training resource level of en-de task. The words in red color are mistranslated and the words in green color are the improvement in transformer + attention link architecture.

In the following Table \ref{tab2}, \ref{tab3}, and \ref{tab4}, we show the BLEU score of the model on the test set under different resource configurations. We train the model on a single P100 GPU for no more than 3 days for each task. Firstly, we see that the transformer with attention link can achieve better results under different training resource settings than the transformer model alone. Secondly, we found that generally the smaller the amount of resources, the greater the improvement. On average, the attention link architecture can improve by 1.0 BLEU score in the 10K training  resource level, improve by 0.8 BLEU score in 30K training  resource level, and improve by 0.3 BLEU score in the full dataset. 
Thirdly, we found that generally the improvements brought by AL  changes with  language. For example, in the 10K training resource level, the improvement on the en-de task is 1.1. But also in the 10K training  resource level, the improvement on the it-en task is 0.6.

In Table \ref{tab5}, we show the training results on bn-gu and gu-ta translation tasks of the CVIT PIB dataset, reflecting that the model brings significant improvements in real low-resource scenarios.

\begin{table*}[]
\caption{Some translation examples of models trained in  10K training resource level of en-de task with.}
\label{n1_tab2}
\begin{center}
    
\begin{tabular}{ll}
\hline
\multicolumn{1}{l|}{source sentence}                & this is hard .                         \\
\multicolumn{1}{l|}{transformer result}             & das ist \color{red}{schwierig} .                    \\
\multicolumn{1}{l|}{transformer+AL result}          & das ist \color{green}{schwer} .                       \\
\multicolumn{1}{l|}{target sentence (ground truth)} & das ist schwer .                       \\ \hline
                                                    &                                        \\ \hline
\multicolumn{1}{l|}{source sentence}                & tremendously exciting .                \\
\multicolumn{1}{l|}{transformer result}             & \color{red}{enorm} .                       \\
\multicolumn{1}{l|}{transformer+AL result}          & \color{red}{sehr} \color{green}{aufregend} .                  \\
\multicolumn{1}{l|}{target sentence (ground truth)} & ungeheuer aufregend .                  \\ \hline
                                                    &                                        \\ \hline
\multicolumn{1}{l|}{source sentence}                & tell me about this world .          \\
\multicolumn{1}{l|}{transformer result}             & \color{red}ich \color{black}erzählen \color{red}ihnen darüber nach .        \\
\multicolumn{1}{l|}{transformer+AL result}          & erzählen \color{green}sie mir \color{red}über diese \color{green}welt .\\
\multicolumn{1}{l|}{target sentence (ground truth)} & erzählen sie mir von dieser welt .     \\ \hline
                                                    &                                        \\ \hline
\multicolumn{1}{l|}{source sentence}                & now , what does that mean ?           \\
\multicolumn{1}{l|}{transformer result}             & \color{red}und \color{black}was bedeutet das ?         \\
\multicolumn{1}{l|}{transformer+AL result}          & \color{green}nun , \color{black}was bedeutet das ? \\
\multicolumn{1}{l|}{target sentence (ground truth)} & nun , was bedeutet das ? \\ \hline
$^{\mathrm{a}}$we show some translation examples of \\models trained in 10K training resource \\level of  en-de task.
The words in red color \\are mistranslated  and the words in
green \\color are the improvement in transformer \\+ attention
link architecture.
\end{tabular}
\end{center}
\end{table*}

\begin{table}[H]
  \caption{BLEU of transformer and transformer with attention link training on the 10K training resource level}
  \begin{center}
  \begin{tabular}{lll}
    \hline
    Task      & transformer & transformer+AL \\
    \hline
    en-de     & 13.3       & \textbf{14.4}   \\
    de-en     & 17.3       & \textbf{18.0}      \\
    en-fr     & 22.9       & \textbf{23.9}    \\
    en-it     & 15.1       & \textbf{16.4}      \\
    it-en     & 19.2       & \textbf{19.8}    \\
    en-ro     & 13.5       & \textbf{14.7}    \\
    Avg       & 16.9       & \textbf{17.9}    \\
    \hline
  \end{tabular}
  \end{center}
  \label{tab2}
\end{table}

\begin{table}[H]
  \caption{BLEU of transformer and transformer with attention link training on 30K  training resource level}
  \begin{center}
  \begin{tabular}{lll}
    \hline
    Task      & transformer  & transformer+AL \\
    \hline
    en-de     & 20.7       & \textbf{21.6}      \\
    de-en     & 25.3       & \textbf{26.1}                 \\
    en-fr     & 31.6       & \textbf{32.3}      \\
    en-it     & 23.4       & \textbf{24.3}       \\
    it-en     & 26.5       & \textbf{27.3}           \\
    en-ro     & 20.9       & \textbf{21.7}    \\
    Avg       & 24.7       & \textbf{25.5}    \\
    \hline
  \end{tabular}
  
  \label{tab3}
  \end{center}
\end{table}

\begin{table}[H]
  \caption{BLEU of transformer and transformer with attention link training on full IWSLT14 training dataset}
  
  \begin{center}
      
  \begin{tabular}{lll}
    \hline
    Task      & transformer  & transformer+AL\\
    \hline
    en-de     & 28.6       & \textbf{28.7}          \\
    de-en     & 34.4       & \textbf{34.7}       \\
    en-fr     & 40.4       & \textbf{40.9}      \\
    en-it     & 30.8       & \textbf{31.2}     \\
    it-en     & 34.7       & \textbf{34.9}      \\
    en-ro     & 28.3       & \textbf{29.0}     \\
    Avg       & 32.9       & \textbf{33.2}    \\
    \hline
  \end{tabular}
  
  \label{tab4}
  \end{center}
\end{table}

\begin{table}[H]
  \caption{BLEU of transformer and transformer with attention link training on CVIT PIB training dataset}
  \begin{center}
  \begin{tabular}{lll}
    \hline
    Task      & transformer  & transformer+AL\\
    \hline
    bn-gu     & 9.1       & \textbf{10.0}          \\
    gu-ta     & 10.3       & \textbf{10.4}      \\
    Avg       & 9.8       & \textbf{10.2}    \\
    \hline
  \end{tabular}
  
  \label{tab5}
  \end{center}
\end{table}

\subsection{Combined with other advanced models }
In this section, we show that our proposed attention link architecture can be easily combined with the current advanced transformer-based NMT models, and can bring performance improvements. We tried to combine attention link with other existing advanced transformer-based models: rdrop \cite{wu2021r} and cutoff \cite{shen2020simple}.
We reached a new sota on the IWSLT de-en task with a 37.9 BLEU score after combining with the cutoff \cite{shen2020simple} model. Since the attention link does not introduce new parameters and hyper-parameters and to ensure fairness, we keep the same hyper-parameters settings after combined in our experiments. The experimental results are shown in Table \ref{tab6}.

\begin{table}[H]
  \caption{BLUE of combining attention link with other transformer-based methods}
  \begin{center}
  \begin{tabular}{lll}
    \hline
    model               & BLEU \\
    \hline
    R-drop\cite{wu2021r}              & 36.9$\pm0.5$                             \\
    cut-off\cite{shen2020simple}             & 37.7$\pm0.5$           \\
    \hline
    R-drop\cite{wu2021r}+AL           & 37.3$\pm0.5$     \\
    cut-off\cite{shen2020simple}+AL          & \textbf{37.9$\pm0.5$}     \\
    \hline
  \end{tabular}
  
  \label{tab6}
  \end{center}
\end{table}

\begin{table}[H]
  
  \begin{center}
    \caption{BLEU of the vanilla transformer (wo AL), only AL in the encoder  (enc AL), only AL in the encoder (dec AL), AL in both encoder and decoder (AL)}
  \begin{tabular}{lllll}
    \hline
    Task      & wo AL & enc AL & dec AL & AL\\
    \hline
    en-de     & 13.3    &13.8  &14.2 & \textbf{14.4}   \\
    de-en     & 17.3    &\textbf{18.1}  &17.7 & 18.0      \\
    en-fr     & 22.9   &\textbf{24.3}  &23.5 & 23.9    \\
    en-it     & 15.1    &15.7  &15.5 & \textbf{16.4}      \\
    it-en     & 19.2    &19.7  &19.2 & \textbf{19.8}    \\
    en-ro     & 13.5    &14.1  &14.3 & \textbf{14.7}    \\
    Avg       & 16.9    &17.6  &17.4 & \textbf{17.9}    \\
    \hline
  \end{tabular}
  
  \label{tab7}
  \end{center}
\end{table}

\section{Ablation study}
To further illustrate the effectiveness of our proposed attention link, we conduct ablation experiments, that is, only add the attention link to the encoder module of the transformer and only add the attention link to the decoder module of the transformer. We perform ablation experiments on the 10K training resource level of six tasks of IWSLT: en-de, de-en, en-fr, en-it, it-en, and en-ro.  

The experimental results are shown in Table \ref{tab7}. We can see that only adding the attention link on the encoder module of the transformer and only adding the attention link on the decoder module of the transformer will generally improve the performance compared to the original transformer model. However, the improvement is generally not as apparent as adding the attention link on both the encoder and decoder sides of the transformer. This further illustrates the effectiveness of the attention link. Besides, we can also see that adding the attention link on the encoder module is generally more effective. 

\section{Conclusion}
In this paper, motivated by shortcut design, we designed a new and efficient architecture named attention link. We theoretically explain the superiority of the attention link. Furthermore, experiments have shown that the transformer model with the attention link achieves much better performance and we achieved a new sota on IWSLT14 de-en translation task with a 37.9 BLEU score by combining  attention link and other advanced method.

We believe that attention links are easy to capture more general semantic information in sentences and are robust to noise. Furthermore, our architecture can be easily deployed in transformer-based models without adding extra parameters. In some sense, it's a free lunch. 
We hope our research can provide a new idea for obtaining an efficient NMT model with low resources.

For limitations, this paper mainly studies the low-resource NMT model. And the attention link we proposed is based on the transformer model, so it may also inherit the shortcomings of the transformer, such as difficulty in selecting suitable optimization parameters.

\bibliographystyle{IEEEtran}
\bibliography{IEEE}

\end{document}